\documentclass[10pt,a4paper,conference]{IEEEtran}
\usepackage[noend]{algpseudocode}
\usepackage{algorithmicx,algorithm}
\usepackage{microtype}
\usepackage{times}
\usepackage{epsfig}
\usepackage{graphicx}
\usepackage{subfigure}
\usepackage{booktabs} 
\usepackage{amsmath}
\usepackage{amssymb}
\usepackage{multirow}

\ifCLASSINFOpdf
\else
\fi
\begin{document}
%
\title{P-DIFF: Learning Classifier with Noisy Labels based on Probability Difference Distributions}

\author{\IEEEauthorblockN{Wei Hu, QiHao Zhao, Fan Zhang}
\IEEEauthorblockA{Dept. of Computer Science and Technology\\
Beijing University of Chemical Technology\\
Beijing, China\\
\{huwei; qhzhao; fzhang\}@mail.buct.edu.cn
}
\and
\IEEEauthorblockN{Yangyu Huang}
\IEEEauthorblockA{Microsoft Research, Asia\\
Beijing, China\\
yangyu.huang@microsoft.com
}}


%


\maketitle

\begin{abstract}
Learning deep neural network (DNN) classifier with noisy labels is a challenging task because the DNN can easily over-fit on these noisy labels due to its high capability. In this paper, we present a very simple but effective training paradigm called \emph{P-DIFF}, which can train DNN classifiers but obviously alleviate the adverse impact of noisy labels. Our proposed probability difference distribution implicitly reflects the probability of a training sample to be clean, then this probability is employed to re-weight the corresponding sample during the training process. P-DIFF can also achieve good performance even without prior-knowledge on the noise rate of training samples. Experiments on benchmark datasets also demonstrate that P-DIFF is superior to the state-of-the-art sample selection methods.
\end{abstract}


%
\IEEEpeerreviewmaketitle

\section{Introduction}
\label{secIntro}
DNN-based classifiers achieve state-of-the-art results in many researching fields. DNNs are typically trained with large-scale carefully annotated datasets. However, such datasets are difficult to obtain for classification tasks with large numbers of classes. Some approaches~\cite{fergus2010learning,divvala2014learning,krause2016unreasonable,niu2015visual} provide the possibility to acquire large-scale datasets, but inevitably result in noisy/incorrect labels, which will adversely affect the prediction performance of the trained DNN classifiers.

To solve this problem, some approaches try to estimate the noise transition matrix to correct mis-labeled samples~\cite{menon2015learning,liu2016classification,natarajan2013learning,han2018masking}. However, this matrix is difficult to be accurately estimated, especially for classification with many classes. Other correction methods~\cite{patrini2017making,ghosh2017robust,li2017learning,ma2018dimensionality} are also proposed to reduce the effect of noisy labels. Recently, some approaches focus on selecting clean samples, and update the DNNs only with these samples~\cite{mentornet,decoupling,coteaching,wang2018iterative,coteaching_plus,incv,o2unet,reweight}.

In this paper, we propose P-DIFF, a novel sample selection paradigm, to learn DNN classifiers with noisy labels. Compared with previous sample selection approaches, P-DIFF provides a stable but very simple method for evaluating sample being noisy or clean. The main results and contributions of the paper are summarized as follows:
\begin{enumerate}
\item We propose the P-DIFF paradigm to learn DNN classifiers with noisy labels. P-DIFF uses a probability difference strategy, instead of the broadly utilized small-loss strategy, to estimate the probability of a sample to be noisy. Moreover, P-DIFF employs a global probability distribution generated by accumulating samples of some recent mini-batches, so it demonstrates more stable performance than single mini-batch approaches. P-DIFF paradigm does not depend on extra datasets, phases, models or information, and is very simple to be integrated into existing softmax-loss based classification models.
\item Compared with SOTA sample selection approaches, P-DIFF has advantages in many aspects, including classification performance, resource consumption, computation complexity. Experiments on several benchmark datasets, including a large real-world noisy dataset cloth1M~\cite{cloth1m}, demonstrate that P-DIFF outperforms previous state-of-the-art sample selection approaches at different noise rates.
\end{enumerate}

\section{Related Work}
\label{secRelated}
Learning with noisy datasets has been widely explored in classification~\cite{frenay2014classification}. Many approaches use pre-defined knowledge to learn the mapping between noisy and clean labels, and focus on estimating the noise transition matrix to remove or correct mis-labeled samples~\cite{menon2015learning,liu2016classification,natarajan2013learning}.

Recently, it has also been studied in the context of DNNs. DNN-based methods to estimate the noise transition matrix are proposed too~\cite{xiao2015learning,sukhbaatar2014training,goldberger2016training,veit2017learning,hendrycks2018using} use a small clean dataset to learn a mapping between noisy and clean annotations. \cite{patrini2017making,ghosh2017robust} use noise-tolerant losses to correct noisy labels. \cite{li2017learning} constructs a knowledge graph to guide the learning process. \cite{han2018masking} proposes a human-assisted approach which incorporates an structure prior to derive a structure-aware probabilistic model. Local Intrinsic Dimensionality is employed in ~\cite{ma2018dimensionality} to adjust the incorrect labels. Rank Pruning~\cite{northcutt2017rankpruning} is proposed to train models with confident samples, and it can also estimate noise rates. However, Rank Pruning only aims at binary classification. \cite{jointoptimization} implements a simple joint optimization framework to learn the probable correct labels of training samples, and to use the corrected labels to train models. \cite{selflearning} proposes an extra Label Correction Phase to correct the wrong labels and achieve good performance. Yao et al.~\cite{lccn} employ label regression for noisy supervision.

Some approaches attempt to update the DNNs only with separated clean samples, instead of correcting the noisy labels. A Decoupling technique~\cite{decoupling} trains two DNN models to select samples that have different predictions from these two models. Weighting training samples~\cite{friedman2001elements,focalloss,selfpaced} is also applied to select clean samples. Based on \emph{Curriculum learning}~\cite{curriculum}, some recent proposed approaches select clean training samples by using some strategies. However, these approaches usually require extra consumption, such as reference or clean sets~\cite{reweight}, extra models~\cite{mentornet,learningtolearn,cleannet,coteaching,coteaching_plus,metaweight}, or iterative/multi-step training~\cite{wang2018iterative,o2unet}.

In the paper, we proposed a very simple sample selection approach P-DIFF. Compared with previous approaches, \emph{reference sets}, \emph{extra models}, and \emph{iterative/multi-step training} are \textbf{not} required in P-DIFF.

\section{The proposed P-DIFF Paradigm}
\label{secPDIFF}

Samples with incorrect label are referred as \emph{noisy samples}, and their labels are referred as \emph{noisy labels}. Noisy labels fall into two types: \textbf{label flips} where the sample has been given a label of another class within the dataset, and \textbf{outliers}, where the sample does not belong to any of the classes in the dataset. In some papers, they are named as \emph{closed-set} and \emph{open-set} noisy labels. As most of previous works~\cite{mentornet,coteaching,li2017learning,patrini2017making,vahdat2017toward,han2018masking,ma2018dimensionality,o2unet,coteaching_plus}, we also address the noisy label problem in a \textbf{closed-set setting}. Actually, experiments in Section~\ref{secNoTau} on the large real-world dataset Cloth1M~\cite{cloth1m} demonstrate that P-DIFF is capable to process open-set noisy labels too.

Our P-DIFF is also a sample selection paradigm. The key of selecting samples is an effective method to measure the possibility that a sample is clean. Some recently proposed methods~\cite{mentornet,coteaching,reweight,coteaching_plus} employ the small-loss strategy to select clean samples. Different from those approaches, P-DIFF selects the clean samples based on the \textbf{probability difference distributions}. The probability difference distribution is computed with the output of the softmax function, and is presented as follows.

\subsection{Probability Difference Distributions}
\label{secPDD}

\begin{figure}[tbh]
\centering
\subfigure[The 1-st Epoch]
{{\includegraphics[width=0.4\columnwidth]{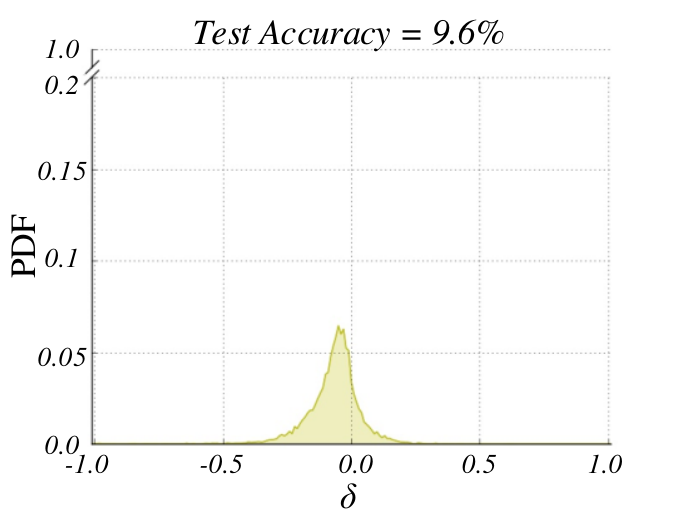}}}
\subfigure[The 2-nd Epoch]
{{\includegraphics[width=0.4\columnwidth]{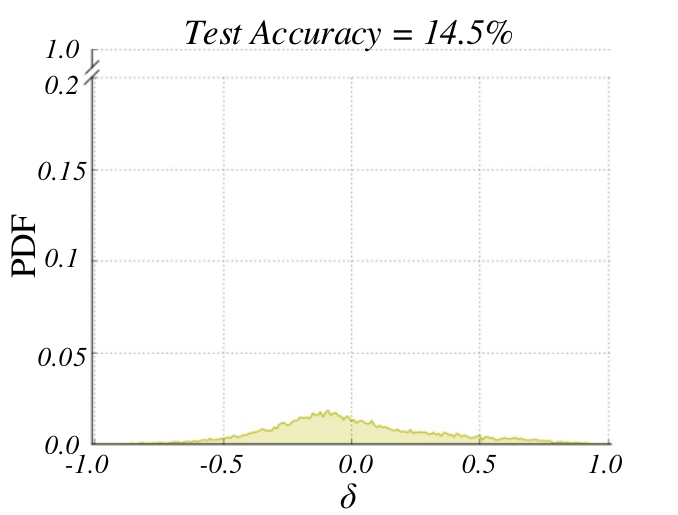}}}
\subfigure[The 10-th Epoch]
{{\includegraphics[width=0.4\columnwidth]{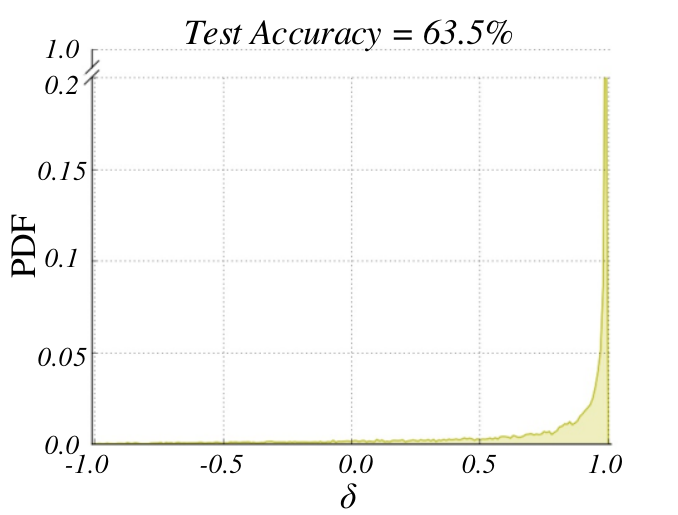}}}
\subfigure[The 200-th Epoch]
{{\includegraphics[width=0.4\columnwidth]{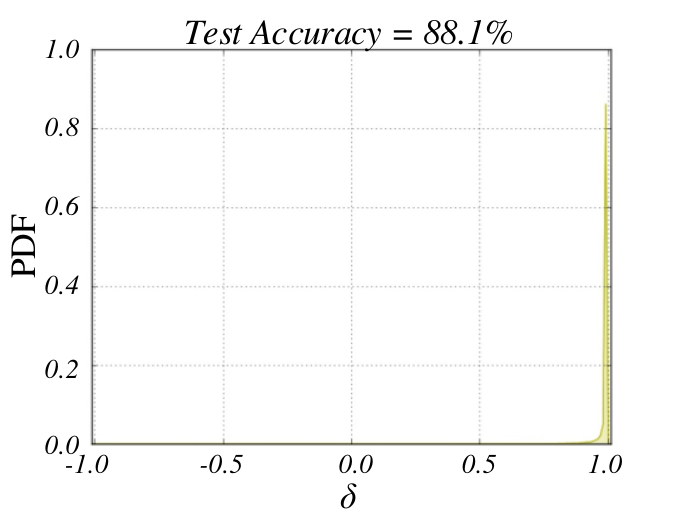}}}
\caption{$DIST_{all}$ and the corresponding performance results at different training epochs.}
\label{figDist1}
\end{figure}

Following previous sample selection approaches, such as O2UNet~\cite{o2unet}, we also remove potential noisy labels to achieve better performance, although noisy and real hard samples may not be distinguished in some cases. Our sample selection strategy is based on two key strategies: \emph{Probability Difference} and \emph{Global Distribution}.

\subsubsection{Probability Difference}
\label{secProbDiff}
The softmax loss is widely applied to supervise DNN classification, and can be considered as the combination of a softmax function and a cross-entropy loss. The output of the softmax function is $\vec{P} = (p_0, p_1, \ldots , p_C)$, where $C$ is the class number. As for an input sample, $p_m \in[0,1]$ is the predicted probability of belonging to the $m$-th class, and

\begin{equation}\label{equSoftmax}
  p_m = \frac{e^{\vec{W}_{m}^T\vec{x}+\vec{b}_{m}}}{\sum_{j=1}^{C}{e^{\vec{W}_{j}^T\vec{x}+\vec{b}_{j}}}},
\end{equation}

where $\vec{x}$ denotes the feature of the input sample computed with DNNs. $\vec{W}$ and $\vec{b}$ are the weight and the bias term in the softmax layer respectively. The cross-entropy loss is defined as

\begin{equation}\label{equCELoss}
  \mathcal{L}=-\sum\limits_{m=1}^C q_mlog(p_m),
\end{equation}
where $q_m$ is the ground truth distribution defined as
\begin{equation}\label{equQK}
q_m=
\begin{cases}
0& m \neq y\\
1& m = y
\end{cases},
\end{equation}
where $y$ is the ground truth class label of the input sample.

Generally, as training a DNN classifier, $p_y$ is encouraged to be the largest component in $\vec{P}$ for an input sample belonging to the $y$-th class. However, if the sample is wrongly labeled, enlarging $p_y$ would lead to adverse effects on the robustness of the trained classifier. The small-loss strategy has been proven to be an effective way to select clean samples ~\cite{mentornet,coteaching,reweight,coteaching_plus}. However, the small-loss strategy cannot select appropriate clean samples in some cases. For example, $\vec{P}_1 = \{0.2, 0.2, 0.2, 0.2, 0.2\}$ and $\vec{P}_2 = \{0.0, 0.2, 0.0, 0.0, 0.8\}$ are the output values of two training samples, and $y = 1$. It is clear that the $\mathcal{L}$ values of these two samples are equal because of the same $p_y = 0.2$, but the second sample has much higher probability to be wrongly labeled.

We define the \textbf{probability difference} $\delta$ of a sample, which belongs to the $y$-th class, as
\begin{equation}\label{equDelta}
  \delta = p_y - p_n,
\end{equation}
where $p_n$ is the largest component except $p_y$ in $\vec{P}$, so $\delta \in [-1,1]$. Ideally, the $\delta$ value should be 1 for a clean sample. If the sample is a label flip noisy sample, we can also ideally infer that $p_y=0$ and $p_n=1$ ($\delta=-1$), where $n$ is the correct label. Although we cannot achieve such results in real training, it inspires us to select clean samples according to $\delta$ values.

It is clear that $\delta_1 = 0.0$ and $\delta_2 = -0.6$ of two samples mentioned above, which indicates that the second sample has higher probability to be noisy. Experiments in Section~\ref{secExp_Delta} verify the effectiveness of $\delta$, compared with only $p_y$.

\subsubsection{Global Distribution}
\label{secGlobal}
Furthermore, only considering samples in one mini-batch~\cite{mentornet,coteaching,reweight} reduces the stabilization of sample selection, and a global threshold is not applied too since the loss values are rapidly changed especially in early epoches. P-DIFF adopts a selection method based on a $\delta$ histogram. We compute the histogram distribution of $\delta$ for all input samples, and this global distribution, called $DIST_{all}$, is just the \textbf{probability difference distribution}. We divide the entire range $[-1,1]$ of the distribution into $H$ bins. We set $H=200$ in our implementation. Let $PDF(x)$ be the ratio of samples whose $\delta$ fall into the $x$-th bin as
\begin{equation}\label{equPDF}
  PDF(x) = \frac{1}{N} \sum_{i=1}^{N}
\begin{cases}
1& \lceil H\cdot\frac{\delta_i+1}{2} \rceil = x\\
0& else
\end{cases},
\end{equation}
where $N$ is the number of training samples. $PDF(x)$ means the probability distribution function of $DIST_{all}$. We then define the probability cumulative function of $DIST_{all}$ as
\begin{equation}\label{equPCF}
  PCF(x) = \sum_{i=1}^{x}PDF(i).
\end{equation}

Moreover, given the $x$-th bin, we can get its value range as
\begin{equation}\label{equRange}
  \delta \in (2\cdot\frac{x-1}{H}-1, 2\cdot\frac{x}{H}-1].
\end{equation}

We perform an experiment to show this distribution. The experiment setting is presented in the Section~\ref{secExp}. We train a normal DNN model with Cifar-10. Figure~\ref{figDist1} shows a probability difference distribution of the DNN at different training epochs. This distribution $DIST_{all}$ is employed in our P-DIFF paradigm to learn classifier with noisy labels. In theory, the distribution $DIST_{all}$ should be computed in each mini-batch training, but it is time-consuming if the number of samples is large. In our implementation, only the $\delta$ values of samples belonging to recent $M$ mini-batch samples are stored to generate the distribution $DIST_{sub}$. If $M$ is too small, $DIST_{sub}$ cannot be considered as a good approximation of $DIST_{all}$. However, a large $M$ is not appropriate too, because the $\delta$ values of far earlier training samples cannot approximate their current values (discussed in Section~\ref{secExp_M}).

\subsection{Learning Classifier with Noisy Labels}

The basic idea of P-DIFF paradigm is trying to select clean samples based on the $DIST_{all}$. As discussed in the Section~\ref{secPDD}, the samples with larger $\delta$ have higher probability of being clean during training, and they should have higher rate to be selected to update the training DNN model. The remaining problem is to find a suitable threshold $\hat{\delta}$.

Given a noise rate $\tau$ and a distribution $DIST_{all}$, P-DIFF drops a certain rate ($\tau$) of the training samples that fall into the left part of $DIST_{all}$. We simply find the smallest bin number $x$ which makes
\begin{equation}\label{equSel1}
  PCF(x) > \tau.
\end{equation}
Therefore, all samples falling into left of the $x$-th bin will be dropped in training. According to Equation~\ref{equRange}, the $\delta$ values of these samples should be less than $2\cdot(x-1)/H-1$, and we can define the threshold $\hat{\delta}$ as
 \begin{equation}\label{equThresh}
   \hat{\delta} =  2\cdot\frac{x-1}{H}-1.
 \end{equation}

However, at the beginning of training process, the DNNs do not have the ability to classify samples correctly, so we cannot drop training samples with the rate $\tau$ throughout the whole training process. We know the DNNs will be improved as the training iteration increases. Therefore, similar with Co-teaching~\cite{coteaching}, we define a dynamic drop rate $R(T)$, where $T$ is the number of training epoch, as
\begin{equation}\label{equRT}
  R(T)=\tau\cdot \min(\frac{T}{T_k},1).
\end{equation}
We can see that all samples are selected at the beginning, then more and more samples are dropped as $T$ gets larger until $T=T_k$ (a given epoch number), and the final drop rate is $\tau$. Therefore, Equation~\ref{equSel1} is re-written as
\begin{equation}\label{equSel2}
  PCF(x) > R(T).
\end{equation}

P-DIFF updates DNN models by redefining Equation~\ref{equCELoss} as
\begin{equation}\label{equWCELoss}
  \mathcal{L}=-\omega \sum\limits_{m=1}^C q_mlog(p_m),
\end{equation}
where $\omega$ is the computed weight of the sample. We set $\omega=1$ if $\delta > \hat{\delta}$, or $\omega$ is set to 0.

Algorithm~\ref{algPDIFF} gives the detailed implementation of our P-DIFF paradigm with a given noise rate $\tau$.

\begin{algorithm}[tb]
   \caption{P-DIFF Paradigm}
   \label{algPDIFF}
\begin{algorithmic}
   \State {\bfseries Input:} Training Dataset $D$, epoch $T_k$ and $T_{max}$, iteration per-epoch $Iter_{epoch}$, batch size $S_{batch}$, noise rate $\tau$, batch rate $M$;
   \State {\bfseries Output:} DNN parameter $\vec{W}$;
   \State
   \State Initialize $\vec{W}$;
   \For{$T=1$ {\bfseries to} $T_{max}$}
   \State Compute the rate $R(T)$ using Equation~\ref{equRT};
   \For{$Iter=1$ {\bfseries to} $Iter_{epoch}$}
   \State Compute the threshold $\hat{\delta}$ using Equation~\ref{equThresh} and Equation ~\ref{equSel2};
   \State Get the mini-batch $\bar{D}$ from $D$;
   \State Set the gradient $G=0$;
   \For{$S=1$ {\bfseries to} $S_{batch}$}
   \State Get the $S$-th sample $\bar{D}(S)$;
   \State Compute $\vec{P}$ of $\bar{D}(S)$ using $\vec{W}$;
   \State Compute the $\delta$ value using Equation~\ref{equDelta};
   \If{$\delta > \hat{\delta}$}
   \State $\omega=1$;
   \Else
   \State $\omega=0$;
   \EndIf
   \State $G += \nabla \mathcal{L}$ (see Equation~\ref{equWCELoss});
   \EndFor
   \State Update $DIST_{sub}$ with the computed $\delta$ values of the last $M\times Iter_{epoch}$ mini-batches;
   \State Update the parameter $\vec{W}=\vec{W}-\eta \cdot G$;
   \EndFor
   \EndFor
\end{algorithmic}
\end{algorithm}

\subsection{Training without a given $\tau$}
\label{secNoTau}
Similar with Co-teaching, a given noise rate $\tau$ is required to compute $R(T)$ in P-DIFF (Algorithm~\ref{algPDIFF}). If $\tau$ is not known in advance, it can be inferred by using the validation set as~\cite{coteaching,liu2016classification}. However, the rate inferred using the validation set cannot always accurately reflect the real rate in the training set. We further explore the method for learning classifiers without a pre-given noise rate $\tau$.

According to the algorithm described above, the key of P-DIFF is to find a suitable threshold $\hat{\delta}$ to separate clean and noisy training samples. Based on the definition $\delta = p_y - p_n$, we can reasonably infer that 0 might be a candidate. Considering the gradually learning problem (see Equation~\ref{equRT}), we compute the threshold as
\begin{equation}\label{equHat}
  \hat{\delta} = \min(\frac{T}{T_k},1) - 1.
\end{equation}
In other words, all samples are employed to learn the classifier in the beginning, and with the increase of the training epoch number, some samples will be dropped. At last, only samples with $\delta>0$ are selected as clean samples to update the DNNs.

\subsubsection{Noise Rate Estimation} DNNs memorize easy/clean samples firstly, and gradually adapt to hard/noisy samples as training epochs become large~\cite{arpit2017closer}. When noisy label exists, DNNs will eventually memorize incorrect labels. This phenomenon, called Noise-Adaption phenomenon, does not change with the choice of training optimizations or networks. DNNs can memorize noisy labels, so we cannot only trust $\hat{\delta}=0$. In the section, we further propose a noise rate estimation technique to achieve better performance.

According to the definition of $\delta$, the $\delta$ value should be encouraged to be close to 1 or -1 for clean and noisy samples respectively. Therefore we propose a value $\zeta$ to evaluate the performance of the learned classifier as
\begin{equation}\label{equEva}
  \zeta = \sum_{x=1}^{H}(|2\cdot \frac{x-1}{H}-1|\cdot PDF(x)).
\end{equation}
In fact, $\zeta \in [0,1]$ is the expected value of $|\delta|$ in the distribution $DIST_{all}$. According to the Noise-Adaption phenomenon and the P-DIFF paradigm, a high $\zeta$ should indicate that the DNN model currently mainly memorizes correct labels. Therefore, we can reasonably infer that the proportion of noisy samples in all samples with $\delta>\hat{\delta}$ is small, and the noise rate $\tau$ can be estimated based on the above inference.

We firstly train the DNN model until $T=T_k$ ($\hat{\delta}=0$), then $\zeta$ is computed for each mini-batch. Once $\zeta$ is larger than a threshold (discussed in Section~\ref{secExp_Zeta}), all samples with $\delta<0$ are regarded as noisy samples to estimate the noise rate $\tau$. With the estimated $\tau$, the threshold $\hat{\delta}$ is then computed by using Equation~\ref{equThresh} instead of Equation~\ref{equHat}, and we train DNNs by using the method with the given noise rate $\tau$ as Algorithm~\ref{algPDIFF}. If $\zeta$ is always less than the threshold, we estimate $\tau$ in the end of training by regarding all samples with $\delta<0$ as noisy samples.

\section{Experiments}
\label{secExp}

We verify the effectiveness of P-DIFF on 4 benchmark datasets: MNIST, Cifar10, Cifar100, and Mini-ImageNet~\cite{miniimage}, which are popularly used for evaluation of noisy labels in previous works. Furthermore, we also perform experiments on a large real-world noisy benchmark Cloth1M~\cite{cloth1m}. For the fair comparison, we use 9-layer~\cite{coteaching} and ResNet-101 CNNs in experiments. All models are trained by using the SGD optimizer(momentum=0.9) with an initial learning rate 0.001 on a TitanX GPU. The batch size is set to 128. We fix $T_{max}=200$ to train all CNN classifiers, and fix $T_{k}=20$ in our P-DIFF implementations. Caffe~\cite{caffe} is employed to implement P-DIFF. Following other approaches, we corrupt datasets with two types of noise transition matrix: \textbf{Symmetry flipping} and \textbf{Pair flipping}. 


\begin{figure}[!tbh]
\centering
\subfigure[The 1-st Epoch]
{{\includegraphics[width=0.4\columnwidth]{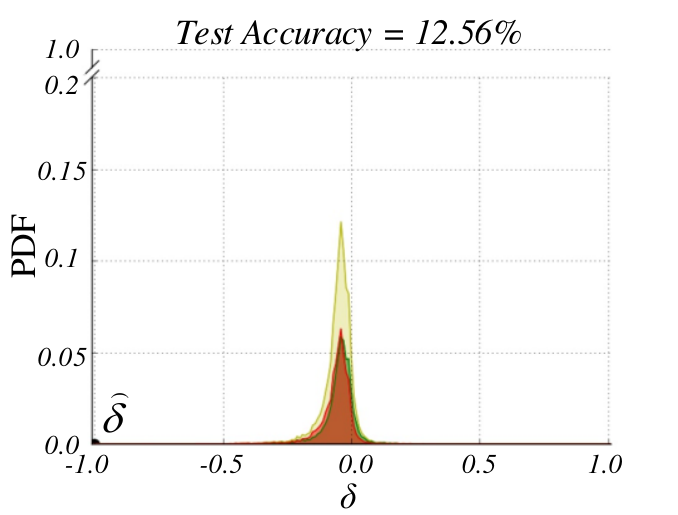}}}
\subfigure[The 8-th Epoch]
{{\includegraphics[width=0.4\columnwidth]{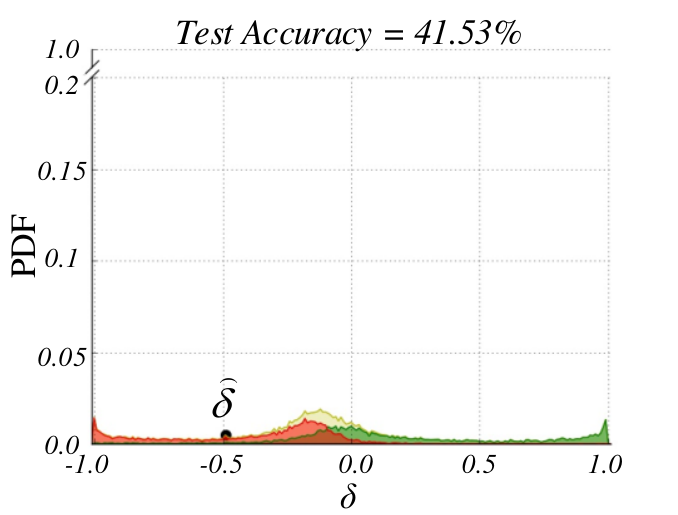}}}
\subfigure[The 21-th Epoch]
{{\includegraphics[width=0.4\columnwidth]{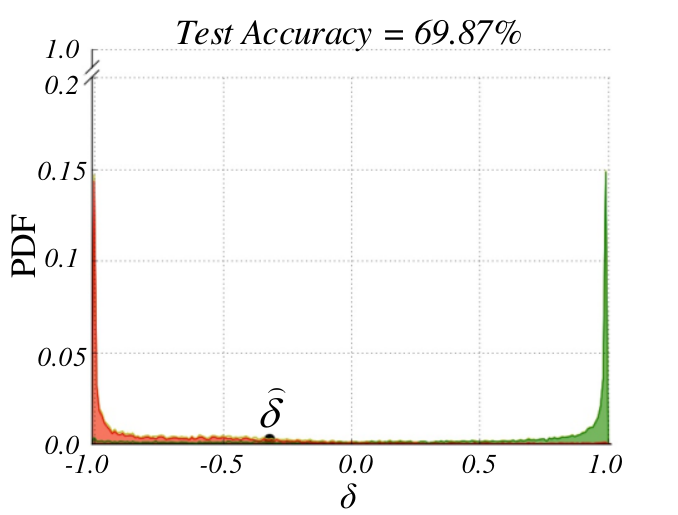}}}
\subfigure[The 200-th Epoch]
{{\includegraphics[width=0.4\columnwidth]{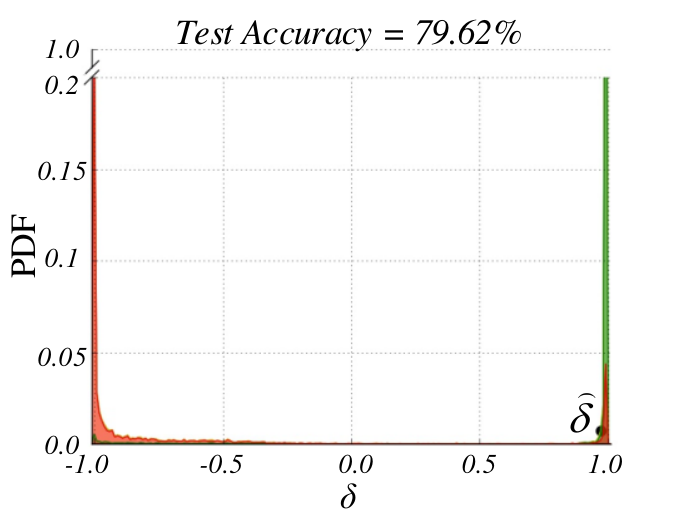}}}
\caption{$DIST_{all}$ (Yellow), $DIST_{clean}$ (Green) and $DIST_{noise}$ (Red) at different training epochs. The DNNs are trained with given noise rates. The corresponding thresholds $\hat{\delta}$ and the performance results can also be seen in the figure.}
\label{figDist2}
\end{figure}

\subsection{Probability Difference Distribution in Training}

We firstly perform an experiment to show the probability difference distribution throughout the training process. In the experiment, Cifar-10 is corrupted by using \textbf{Symmetry flipping} with 50\% noisy rate. To better illustrate the effectiveness of P-DIFF, as shown in Figure~\ref{figDist2}, we present three types of distributions: $DIST_{all}$, $DIST_{clean}$ and $DIST_{noise}$. These distributions are built by using $\delta$ values of all samples, clean samples, and noisy samples respectively. Therefore, we can conclude that $DIST_{all}=DIST_{clean}+DIST_{noise}$.

\begin{figure}[!tbh]
\centering
\subfigure[The 1-st Epoch. $\zeta=0.05$]
{{\includegraphics[width=0.42\columnwidth]{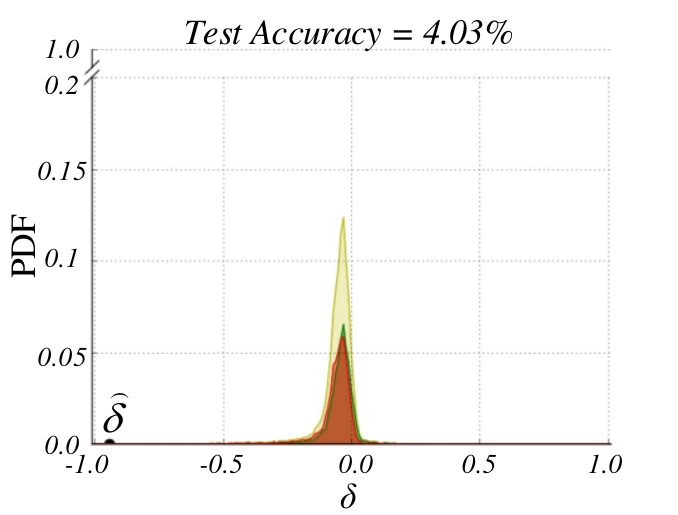}}}
\subfigure[The 8-th Epoch. $\zeta=0.21$]
{{\includegraphics[width=0.42\columnwidth]{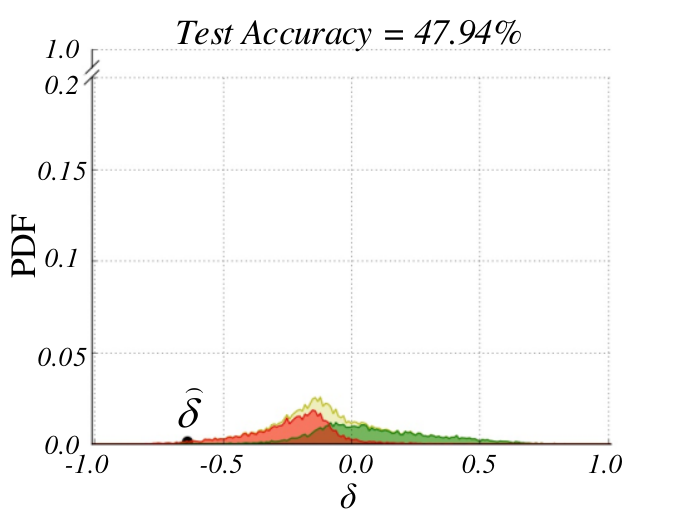}}}
\subfigure[The 21-th Epoch. $\zeta=0.84$]
{{\includegraphics[width=0.42\columnwidth]{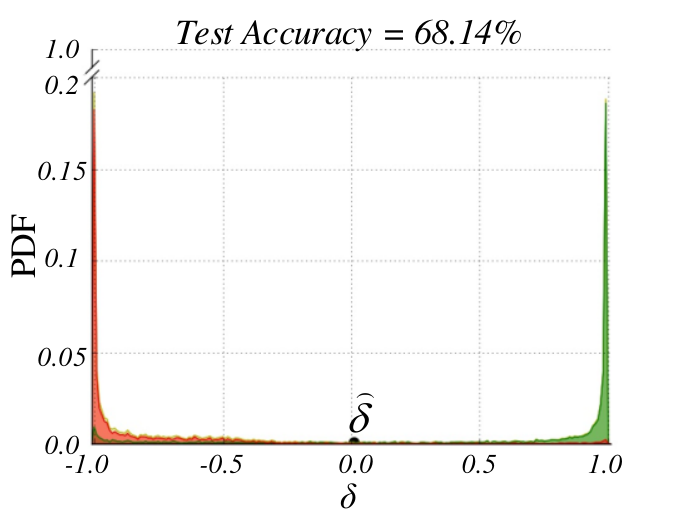}}}
\subfigure[The 200-th Epoch. $\zeta=0.93$]
{{\includegraphics[width=0.42\columnwidth]{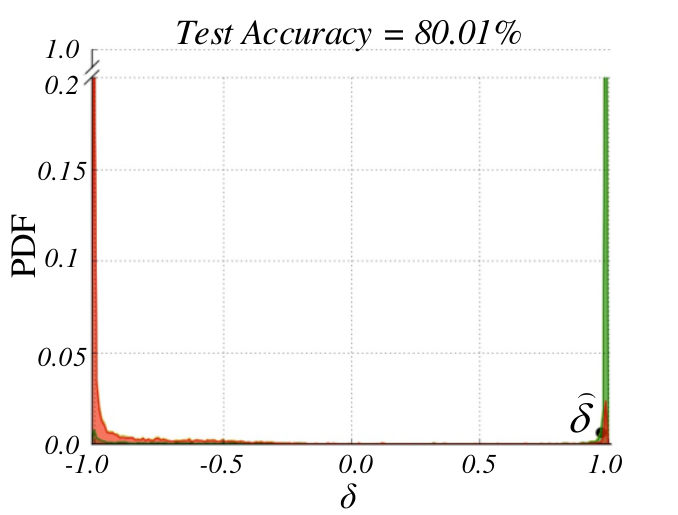}}}
\caption{$DIST_{all}$ (Yellow), $DIST_{clean}$ (Green) and $DIST_{noise}$ (Red) at different training epochs. The DNNs are trained \textbf{without} given noise rates. The corresponding thresholds $\hat{\delta}$, $\zeta$, and the performance results are presented.}
\label{figDist3}
\end{figure}

To evaluate P-DIFF without given noise rates, we perform another experiment on the same noisy dataset, but train the DNN by using the method presented in Section~\ref{secNoTau}. $DIST_{all}$, $DIST_{clean}$ and $DIST_{noise}$ are also presented in Figure~\ref{figDist3}. The Figure~\ref{figDist3} shows $\hat{\delta}$, $\zeta$, and the performance results too. We can see that most of clean and noisy samples are separated clearly by using P-DIFF.

\subsection{Effect of $\delta$}
\label{secExp_Delta}
\begin{figure}[!tbh]
\centering
\subfigure[Pair,45\%]
{{\includegraphics[width=0.3\columnwidth]{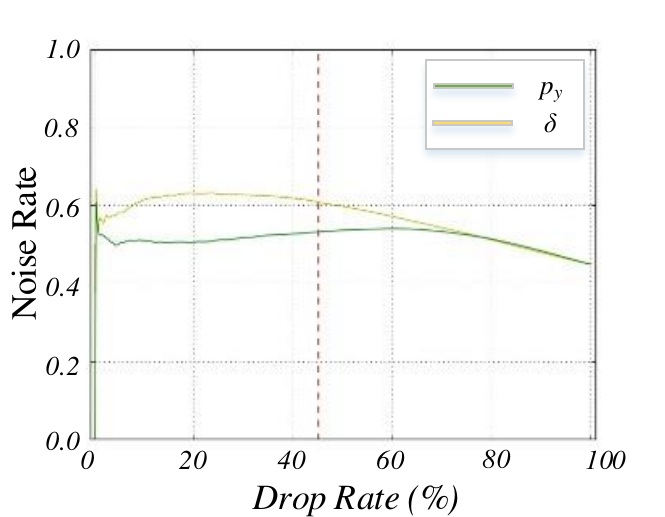}}\label{fig1a}}
\subfigure[Symmetry,20\%]
{{\includegraphics[width=0.3\columnwidth]{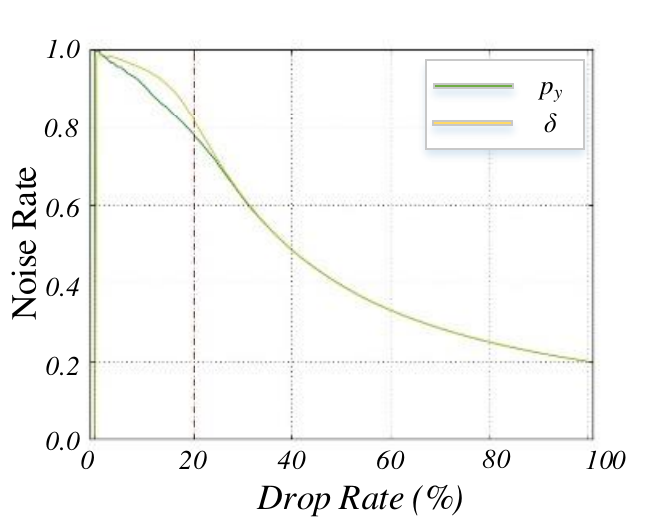}}\label{fig1b}}
\subfigure[Symmetry,50\%]
{{\includegraphics[width=0.3\columnwidth]{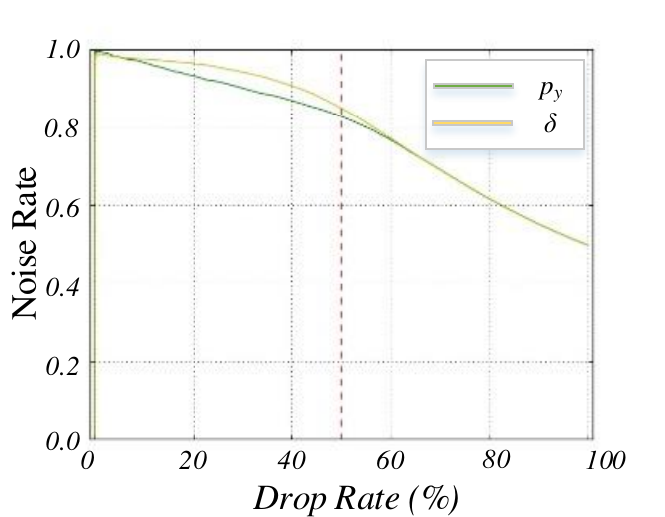}}\label{fig1c}}
\caption{The plotted curves show the relationship between the drop rates and the real noise rates of dropped samples. The green and yellow curves are plotted with the real noise rates computed with two $DIST_{all}$ distributions, which are constructed by employing $p_y$ and $\delta$ respectively. \textbf{Cifar-10} is used in these experiments.}
\label{figCurve}
\end{figure}

We firstly train the DNN models with classical softmax losses on Cifar10. At the 1st iteration of the 2nd epoch, we compute two $DIST_{all}$ distributions constructed with $\delta$ and $p_y$ respectively. We plot two curves to present the relationship between the drop rates and the real noise rates of dropped samples with two distributions, as shown in Figure~\ref{figCurve}. We can observe that the yellow curves are always not lower than the corresponding green curves, which means that more samples with incorrect labels would be dropped if we employ $\delta$ to construct the distribution $DIST_{all}$, especially for the hard \textbf{Pair} noise type. Therefore, using $\delta$ can select more clean samples and should achieve better performance. This phenomenon can also been verified by the following experiment.

To further demonstrate the effectiveness of $\delta$, we train the DNNs (abbreviated as P-DIFF) with some noisy datasets as shown in Table~\ref{tabDelta}. We also employ P-DIFF paradigm to train the DNN (abbreviated as P-DIFF$_{m1}$) but using $p_y$ instead of $\delta$. Comparing the performance of P-DIFF and P-DIFF$_{m1}$ in Table~\ref{tabDelta}, we can see that the probability difference $\delta$ plays the key role to achieve satisfied performance, especially on Pair Flipping datasets.

\begin{table}[!htb]\footnotesize
\caption{Average test accuracy on four testing datasets over the last 10 epochs. P-DIFF$_{m1}$ employs $p_y$ to build the distributions.}
\label{tabDelta}
\begin{center}
\begin{tabular}{lc|cr}
\toprule
DataSet                  & Noise Type, Rate & P-DIFF$_{m1}$ & P-DIFF   \\
\midrule
\multirow{3}{*}{Cifar-10}&Symmetry,20\%  & 85.59\%     & \textbf{88.61}\%    \\
                         &Symmetry,40\%  & 82.74\%     & \textbf{85.31}\%    \\
                         &Pair,10\%      & 83.69\%     & \textbf{87.78}\%    \\
                         &Pair,45\%      & 73.47\%     & \textbf{83.25}\%    \\
\bottomrule
\end{tabular}
\end{center}
\end{table}

\subsection{Effect of $M$}
\label{secExp_M}
$M$ indicates the rate of recent batch number used to generate the distribution $DIST_{sub}$. To demonstrate the effect of $M$, we perform some experiments on \textbf{Cifar-10} with different $M$ value. Table~\ref{tabM} gives the comparison result. We can observe that only using samples in one mini-batch (as ~\cite{mentornet,coteaching,coteaching_plus}) cannot achieve satisfied performance. Meanwhile, a large $M$ is also not preferred as discussed in Section~\ref{secGlobal}, which can be observed from the table too. According to our experiments on several datasets, setting $M=20\%$ can achieve good results in all experiments. Actually, we can observe that $M$ is not a very sensitive parameter for achieving good performance.


\begin{table}[!htb]\footnotesize
\caption{Average test accuracy on four Cifar-10 testing datasets over the last 10 epochs with different $M$. $M=0\%$ means only samples in the single current mini-batch are used.}
\label{tabM}
\begin{center}
\begin{tabular}{c|cccc}
\toprule
$M$ & {\shortstack{Symmetry\\20\%}} & {\shortstack{Symmetry\\40\%}} & {\shortstack{Pair\\10\%}} & {\shortstack{Pair\\45\%}} \\
\midrule
0\%   & 87.71\% & 81.37\% & 84.87\% & 74.23\% \\
5\%   & 88.35\% & 83.09\% & 86.32\% & 77.95\% \\
10\%  & \textbf{88.79}\% & \textbf{85.64}\%  & 88.28\%  &  81.27\% \\
20\%  & 88.61\% & 85.31\%  &\textbf{87.78}\%  & \textbf{83.25}\% \\
50\%  & 88.13\% & 84.14\%  &87.34\%  & 78.04\%\\
100\% & 88.38\% & 85.13\% & 87.67\% & 78.48\%\\
\bottomrule
\end{tabular}
\end{center}
\end{table}

\subsection{Effect of $\zeta$}
\label{secExp_Zeta}
We also perform several experiments to evaluate the effect of $\zeta$ in Equation~\ref{equEva} when the noise rate is not given. $\zeta$ is employed in P-DIFF to reflect the degree of convergence of the model, which can be observed in Figure~\ref{figDist3}. According to the results shown in Table~\ref{tabZeta}, $\zeta$ is not a very sensitive parameter for achieving good performance too, as long as the value is not close to 1.0. Therefore, we set $\zeta=0.9$ in all our experiments.


\begin{table}[!htb]\footnotesize
\caption{Average test accuracy on four Cifar-10 testing datasets over the last 10 epochs with different $\zeta$ values.}
\label{tabZeta}
\begin{center}
\begin{tabular}{c|cccc}
\toprule
$\zeta$ & {\shortstack{Symmetry\\20\%}} & {\shortstack{Symmetry\\40\%}} & {\shortstack{Pair\\10\%}} & {\shortstack{Pair\\45\%}} \\
\midrule
0.5  & 87.71\% & 78.35\% & 82.39\% & 81.49\% \\
0.8  & \textbf{88.28}\% & 84.27\% & 85.38\% & 83.86\% \\
0.85 & 87.37\% & 84.93\%  & 85.65\%  &  86.24\% \\
0.90 & 87.61\% & \textbf{85.74}\%  &\textbf{87.43}\%  & \textbf{86.73}\% \\
0.95 & 86.22\%  &84.49\%  &86.94\%  & 63.24\%\\
1.0 & 86.19\% & 84.13\% & 86.32\% & 60.57\%\\
\bottomrule
\end{tabular}
\end{center}
\end{table}

\subsection{Experiments without a Given $\tau$}
To evaluate P-DIFF without a given noise rate $\tau$, we train the DNNs on benchmarks again, but by using the method presented in Section~\ref{secNoTau}. Moreover, we apply P-DIFF to train DNNs with clean training datasets to demonstrate its effectiveness. The results are shown in Table~\ref{tabNoTau}. From the table, we can see that our estimated $\tau_{est}$ are very close to the real rates in many cases, especially when the corresponding $\zeta$ value is high. This phenomenon also proves that the $\zeta$ can be applied to evaluate the performance of the DNNs trained with P-DIFF.

To verify the effectiveness of $\zeta$, Table~\ref{tabNoTau} presents the test accuracy results (abbreviated as TA$_1$, and $TA_1 = TA$ if $\zeta<0.9$) without considering $\zeta$ (using Equation~\ref{equHat}). $TA$ should be equal to $TA_1$ if $\zeta$ cannot exceed a threshold 0.9 throughout the training process. As shown in the table, the performance of DNNs can be further improved if the noise rates can be estimated with $\zeta$. We also observed that P-DIFF can deal with clean datasets and achieved good results too.

By comparing with the results in Table~\ref{tabDelta}, it is surprised that the DNNs trained without given noise rates even achieve better performance than the DNNs trained with correct given noise rates. More exploration should be conducted to find the reason behind this phenomenon.

\begin{table}[!htb]\footnotesize
\caption{Test accuracy $TA$, estimated rate $\tau_{est}$ and the corresponding $\zeta$ on three datasets. We supply $TA_1$, the test accuracy of the DNNs trained without considering $\zeta$, for comparison.}
\label{tabNoTau}
\begin{center}
\begin{tabular}{p{1cm}|c|cc|c|p{0.2cm}}
\toprule
DataSet                  & Noise Type, Rate &  $TA_1$ & $TA$ & $\tau_{est}$   & $\zeta$\\
\midrule
\multirow{3}{*}{MNIST}   &Clean, 0\%      & 99.62\% & \textbf{99.68}\%    &  0.2\% & 0.99    \\
                         &Symmetry, 20\%  & 99.53\% & \textbf{99.58}\%    & 20.5\% & 0.99    \\
                         &Symmetry, 40\%  & 99.23\% & \textbf{99.43}\%    & 40.2\% & 0.99    \\
                         &Pair, 10\%      & 99.56\% & \textbf{99.61}\%    & 10.4\% & 0.99    \\
                         &Pair, 45\%      & 98.62\% & \textbf{98.70}\%    & 44.6\% & 0.99    \\
\midrule
\multirow{3}{*}{Cifar-10}&Clean, 0\%      & 90.68\% & \textbf{91.18}\%     &  7.3\% & 0.96    \\
                         &Symmetry, 20\%  & 87.12\% & \textbf{87.61}\%     & 28.6\% & 0.96    \\
                         &Symmetry, 40\%  & 85.31\% & \textbf{85.74}\%     & 42.7\% & 0.93    \\
                         &Pair, 10\%      & 86.82\% & \textbf{87.43}\%     & 10.1\% & 0.96    \\
                         &Pair, 45\%      & 85.76\% & \textbf{86.73}\%     & 45.8\% & 0.95    \\
\midrule
\multirow{3}{*}{Cifar-100}&Clean, 0\%     & - & 64.99\%     & 11.8\% & 0.83    \\
                         &Symmetry, 20\%  & - & 62.87\%     & 29.2\% & 0.85    \\
                         &Symmetry, 40\%  & - & 52.43\%     & 47.3\% & 0.71    \\
                         &Pair, 10\%      & - & 63.26\%     & 12.8\% & 0.84    \\
                         &Pair, 45\%      & - & 43.24\%     & 39.3\% & 0.80    \\
\midrule
\multirow{3}{*}{\shortstack{Mini-\\ImageNet}}&Clean, 0\%     & - & 55.81\%     & 12.4\% & 0.81    \\
                         &Symmetry, 20\%  & - & 53.63\%     & 30.4\% & 0.83    \\
                         &Symmetry, 40\%  & - & 46.34\%     & 48.6\% & 0.69    \\
                         &Pair, 10\%      & - & 54.76\%     & 13.3\% & 0.83    \\
                         &Pair, 45\%      & - & 37.14\%     & 42.3\% & 0.79    \\
\bottomrule
\end{tabular}
\end{center}
\end{table}

%

\subsection{Comparison with State-of-the-art Approaches}

We compare the P-DIFF with four outstanding sample selection approaches: Co-teaching~\cite{coteaching}, Co-teaching+~\cite{coteaching_plus}, INCV~\cite{incv}, and O2U-Net~\cite{o2unet}:

\textbf{Co-teaching}: Co-teaching simultaneously trains two networks for selecting samples. We compare Co-teaching because it is an important sample selection approach.

\textbf{Co-teaching+}: This work is constructed on Co-teaching, and heavily depends on samples selected by small-loss strategy. Therefore, it is suitable to compare with P-DIFF for comparison.

\textbf{INCV}: This recently proposed approach divides noisy datasets and utilizes cross-validation to select clean samples. Moreover, the Co-teaching strategy is also applied in the method.

\textbf{O2U-Net}: This work also compute the probability of a sample to be noisy label by adjusting hyper-parameters of DNNs in training. Multiple training steps are employed in the approach. Its simplicity and effectiveness make it to be a competitive approach for comparison.

As the baseline, we also compare P-DIFF with the DNNs (abbreviated as Normal) trained with the same noisy datasets by using the classic softmax loss. The DNNs (abbreviated as Clean) trained only with clean samples (For example, only 80\% clean samples are used for a Symmetry-20\% noisy dataset) are also presented as the \emph{upper bound}. We corrupt datasets with 80\% noise rate to demonstrate that P-DIFF can deal with extremely noisy datasets. Table~\ref{tabFP} reports the accuracy on the testing sets of four benchmarks. We can see that the DNNs trained with P-DIFF are superior to the DNNs trained with these previous state-of-the-art approaches. 


\begin{table*}[!htb]\footnotesize
\caption{Average test accuracy on three testing datasets over the last 10 epochs. Accuracies of O2U-Net are cited from the original paper~\cite{o2unet}, since its authors do not provide the source codes.}
\label{tabFP}
\begin{center}
\begin{tabular}{lc|cc|ccccr}
\toprule
DataSet                  & Noise Type, Rate & Normal & Clean & Co-teaching & Co-teaching++ & INCV & O2U-Net & P-DIFF   \\
\midrule
\multirow{4}{*}{MNIST}   &Symmetry, 20\%  & 94.05\%& 99.68\%& 97.25\%   & 99.26\% & 97.62\% & -  & \textbf{99.58}\%    \\
                         &Symmetry, 40\%  & 68.13\%& 99.51\%& 92.34\%   & 98.55\% & 94.23\% & -  & \textbf{99.38}\%    \\
                         &Symmetry, 80\%  & 23.61\%& 99.04\%& 81.43\%   & 93.79\% & 92.66\% & -  & \textbf{97.26}\%    \\
                         &Pair, 10\%      & 95.23\%& 99.84\%& 97.76\%   & 99.03\% & 98.73\% & -  & \textbf{99.54}\%    \\
                         &Pair, 45\%      & 56.52\%& 99.59\%& 87.63\%   & 83.57\% & 88.32\% & -  & \textbf{99.33}\%    \\
\midrule
\multirow{4}{*}{Cifar-10}&Symmetry, 20\%  & 76.25\%& 89.10\%& 82.66\%   & 82.84\% & 84.87\% & 85.24\% & \textbf{88.61}\%    \\
                         &Symmetry, 40\%  & 54.37\%& 87.86\%& 77.42\%   & 72.32\% & 74.65\% & 79.64\% & \textbf{85.31}\%    \\
                         &Symmetry, 80\%  & 17.28\%& 80.27\%& 22.60\%   & 18.45\% & 24.62\% & 34.93\% & \textbf{37.02}\%    \\
                         &Pair, 10\%      & 82.32\%& 90.87\%& 85.83\%   & 85.10\% & 86.27\% & \textbf{88.22}\% & 87.78\%    \\
                         &Pair, 45\%      & 49.50\%& 87.41\%& 72.62\%   & 50.46\% & 74.53\% &  -      & \textbf{83.25}\%    \\
\midrule
\multirow{4}{*}{Cifar-100}&Symmetry, 20\% & 47.55\%& 66.37\%& 53.79\%   & 52.46\% & 54.87\% & 60.53\% & \textbf{63.72}\%    \\
                         &Symmetry, 40\%  & 33.32\%& 60.48\%& 46.47\%   & 44.15\% & 48.21\% & 52.47\% & \textbf{54.92}\%    \\
                         &Symmetry, 80\%  & 7.65\% & 35.12\%& 12.23\%   &  9.65\% & 12.94\% & \textbf{20.44}\% & 18.57\%    \\
                         &Pair, 10\%      & 52.94\%& 69.27\%& 57.53\%   & 54.71\% & 58.41\% & 64.50\% & \textbf{67.44}\%    \\
                         &Pair, 45\%      & 25.99\%& 61.29\%& 34.81\%   & 27.53\% & 36.79\% & -       & \textbf{45.36}\%    \\
\midrule
\multirow{4}{*}{Mini-ImageNet}&Symmetry, 20\% & 37.83\%& 58.25\%& 41.47\%   & 40.06\% & 43.12\% & 45.32\% & \textbf{56.71}\%    \\
                         &Symmetry, 40\%  & 26.87\%& 53.88\%& 34.81\%   & 34.62\% & 35.65\% & 38.39\% & \textbf{47.21}\%    \\
                         &Symmetry, 80\%  &  4.11\%& 23.63\%&  6.65\%   &  4.38\% &  6.71\% &  8.47\% & \textbf{11.69}\%    \\
                         &Pair, 10\%      & 43.19\%& 61.64\%& 45.38\%   & 43.24\% & 46.34\% & 50.32\% & \textbf{57.85}\%    \\
                         &Pair, 45\%      & 19.74\%& 57.92\%& 26.76\%   & 26.76\% & 28.57\% & -       & \textbf{37.21}\%    \\
\bottomrule
\end{tabular}
\end{center}
\end{table*}

We further perform experiments on a large-scale real-world dataset Cloth1M, which contains 1M/14k/10k train/val/test images with 14 fashion classes. Table~\ref{tabWebVision} lists the performance results. Though P-DIFF addresses noisy problem in the closed-set setting, it can also achieve good results on real-world open-set noisy labels.

\begin{table}[!htb]\footnotesize
\caption{Comparison on Cloth1M}
\label{tabWebVision}
\begin{center}
\begin{tabular}{l|cr}
\toprule
Method & ResNet-101 & 9-Layer CNN \\
\midrule
Coteaching   & 78.52\% & 68.74\% \\
Coteaching++ & 75.78\% & 69.16\% \\
INCV         & 80.36\% & 69.89\% \\
O2U-Net      & 82.38\% & 75.61\% \\
P-Diff       & \textbf{83.67}\% & \textbf{77.38}\% \\
\bottomrule
\end{tabular}
\end{center}
\end{table}

\subsubsection{Comparison on Computational Efficiency} Compared with these approaches, P-DIFF also has advantages in resource consumption and computational efficiency, since other approaches require extra DNN models or complex computation to achieve good performance. Table~\ref{tabTime} shows the training time of these approaches for comparison. All data are measured with the 9-Layer CNNs trained on Cifar-10 with 40\% symmetry noise rate. Furthermore, P-DIFF only requires an extra small memory to store the distribution, so it costs fewer memory than other noise-free approaches too.

\begin{table}[!htb]\footnotesize
\caption{Training time of different approaches. The time of O2U-Net is not provided because of its closed-source.}
\label{tabTime}
\begin{center}
\begin{tabular}{l|cr}
\toprule
Approach       & In Theory & Real Cost/Epoch\\
\midrule
Normal          & $1\times$ & 64 s \\
Co-teaching     & $\approx2\times$ & 131 s  \\
Co-teaching++   & $\approx2\times$ & 143 s  \\
INCV            & $>3\times$ & 217 s  \\
O2U-Net         & $>3\times$ & -  \\
P-DIFF          & $\approx1\times$ & 71 s  \\
\bottomrule
\end{tabular}
\end{center}
\end{table}

\section{Conclusion}

Based on \emph{probability difference} and \emph{global distribution} schemes, we propose a \textbf{very simple but effective} training paradigm P-DIFF to train DNN classifiers with noisy data. According to our experiments on both synthetic and real-world datasets, we can conclude that P-DIFF can achieve satisfied performance on datasets with different noise type and noise rate. P-DIFF has some parameters, such as $M$ and $\zeta$, but we can conclude that the performance of our paradigm is not sensitive to them according to our experiments. Since P-DIFF only depends on a Softmax layer, it can be easily employed for training DNN classifiers. We also empirically show that P-DIFF outperforms other state-of-the-arts sample selection approaches both on classification performance and computational efficiency. Recently, some noise-tolerant training paradigms~\cite{selflearning,jointoptimization} which employ the label correction strategy to achieve good performance, and we will investigate this strategy in P-DIFF to further improve the performance in the future.






%

\bibliographystyle{IEEEtran}
\bibliography{root}

%
%

\end{document}